\documentclass[conference]{IEEEtran}
\IEEEoverridecommandlockouts
\usepackage{cite}
\usepackage{amsmath,amssymb,amsfonts}
\usepackage{algorithmic}
\usepackage{graphicx}
\usepackage{textcomp}
\usepackage{xcolor}
\usepackage{amsmath}
\usepackage{svg}
\usepackage{subfiles}
\usepackage[colorlinks,urlcolor=blue,linkcolor=blue,citecolor=blue]{hyperref}

\usepackage{color,array}

\usepackage{graphicx}

\usepackage[table]{xcolor}
\def\BibTeX{{\rm B\kern-.05em{\sc i\kern-.025em b}\kern-.08em
    T\kern-.1667em\lower.7ex\hbox{E}\kern-.125emX}}
\begin{document}

\title{AD-Relight: Training-Free Banner Relighting via Illumination Translation with Diffusion Priors} 
\author{
\begin{tabular}{@{}c@{\hspace{2.5cm}}c@{}}

\textbf{Rameshwar Mishra$^{*}$} &
\textbf{Bishshoy Das} \\

\textit{Indraprastha Institute of Information Technology} &
\textit{Dolby Laboratories} \\

Delhi, India &
Bangalore, India \\

rameshwarm@iiitd.ac.in &
bishshoy.das@dolby.com \\[1.5em]

\textbf{A. V. Subramanyam} &
\textbf{Guan-Ming Su} \\

\textit{Indraprastha Institute of Information Technology} &
\textit{Dolby Laboratories} \\

Delhi, India &
Sunnyvale, USA \\

subramanyam@iiitd.ac.in &
guanming.su@dolby.com

\end{tabular}

\thanks{$^{*}$This research work was conducted during Rameshwar Mishra's internship at Dolby Labs}
}
\maketitle
\begin{abstract}
The recent surge in content consumption through streaming services has driven a growing demand for personalized content. Personalized advertisements (ads) play a crucial role in enhancing both user engagement and ad effectiveness. A key aspect of ad personalization involves replacing existing regions in a frame with custom, Photoshop-generated banners. However, existing ad-placement pipelines typically rely on simple geometric warping, ignoring the scene’s underlying lighting conditions. Similarly, state-of-the-art diffusion-based object insertion and relighting models struggle to accurately relight these newly inserted banners, as they are not trained on ad-banner data, and training such a model for ad banners will require millions of images. This highlights the need for an effective relighting framework, which results in a seamless integration of custom banners into the original scene. Motivated by this, we present AD-Relight, a novel, multi-stage training-free framework that adapts a diffusion-based relighting model at test time to relight newly added Photoshop-generated ad banners. Through extensive evaluation, we demonstrate that AD-Relight outperforms both relighting baselines and existing ad-placement methods based on simple warping. User studies further show that participants consistently prefer the outputs of AD-Relight over those of prior approaches.  
\end{abstract}

\begin{IEEEkeywords}
Ad-Relighting, Personalization, Training free framework.
\end{IEEEkeywords}

\section{Introduction}

\begin{figure}[t]
  \centering
  \includegraphics[width=\columnwidth]{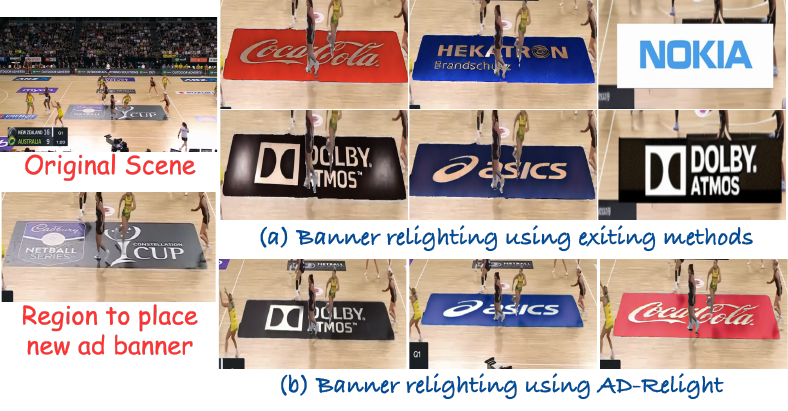}
  \caption{Visual comparison highlighting the limitations of existing methods for banner relighting. In (a), columns 1, 2, and 3 correspond to results from IC-Light \cite{zhang2025scaling}, SpotLight \cite{fortier2024spotlight}, and TF-ICON \cite{lu2023tf}, respectively. All methods fail to reproduce the spatial lighting variations of the scene, leading to inconsistent shading and poor integration of the inserted banner.}
  \label{fig:intro}
  \vspace{-0.6cm}
\end{figure}

With the rapid growth of streaming platforms, there has been a significant shift toward delivering personalized content to viewers. Among various personalization strategies, tailored advertisements have emerged as a key factor in boosting both user engagement and ad performance. Several key components contribute to personalized advertising, including the introduction of new ads, the replacement of existing banners, the selection of optimal ad placements, and the integration of photorealism into added advertisements. In the case of placing a Photoshop-generated custom ad banner in a scene, there exists a key challenge of lighting. As shown in Fig. \ref{fig:intro}(a), we inserted a variety of ad banners using existing relighting methods \cite{fortier2024spotlight, zhang2025scaling} and a training-free object insertion method \cite{lu2023tf}. Across all examples, the inserted banners exhibit lighting inconsistent with the scene. For instance, the original region is more strongly illuminated on the left and far right, creating a gradient that these methods fail to reproduce, leading to unrealistic results. However, as shown in Fig. \ref{fig:intro}(b), AD-Relight produces outputs with banners exhibiting lighting consistent with the original scene.

Several methods have been developed to address ad placement in content. The approach in \cite{efimova2022advertisement} automates banner placement in video frames using a machine learning model that detects banners in the original frames and then inpaints the new ad using a homography matrix. Meanwhile, \cite{akgul2020cloud,li2005real} present end-to-end frameworks for inserting ads into videos. Notably, \cite{akgul2020cloud} introduces a cloud-based approach that places ads based on prior analytics, optimizing system design for faster inference.
Although the previously mentioned methods address ad placement and are closely related to our work, they primarily focus on region estimation and the design of the overall server-side framework. These methods assume that the input ad is already in its final form and only needs to be placed within the scene. However, we found that when banners are placed on the floor, lighting becomes a critical factor in maintaining the realism of the newly introduced ad banners.

With advancements in generative models, such as diffusion models, it is now possible to generate photorealistic content. Several methods have been developed to introduce new objects into images using diffusion models \cite{lu2023tf,song2023objectstitch}. While these diffusion-based methods enhance photorealism, they often struggle to preserve the identity of the added object, a crucial factor for ad banners. Additionally, methods such as \cite{song2023objectstitch} and \cite{song2024imprint} require extensive training on millions of images, making them computationally expensive. \cite{zhang2025scaling}, \cite{jin2024neural}, and \cite{ren2024relightful} propose diffusion-based object relighting frameworks. These models require extensive training on approximately 3 to 5 million images paired with their corresponding light environment maps. While they achieve impressive results in relighting portrait images, they often struggle with banners placed on the floor, as such cases fall outside their training distribution. Fig. \ref{fig:intro} illustrates how direct use of diffusion-based relighting and object insertion methods can lead to inconsistent and unrealistic outputs. Training these models specifically for ad banners would require collecting millions of relevant images, a labor-intensive, challenging, and time-consuming task. 

The following are our observations based on our analysis of current methods. \textcolor{red}{\textit{All the previously mentioned ad placement methods focus on region estimation, cloud-based solutions, and server-side frameworks. Diffusion-based object insertion methods suffer from identity loss and distortion. Diffusion-based relighting models often struggle with banner relighting, particularly for floor-level placements outside their training distribution.}}

To the best of our knowledge, no prior work explicitly tackles the relighting of Photoshop-generated ad banners. We formulate banner relighting as a conditional image-to-image illumination translation problem, where the goal is to transfer the lighting characteristics of the target region onto the inserted banner. In this work, we propose AD-Relight, a framework to relight newly added ad banners in a scene, ensuring they appear consistent with the original scene.

Our main contributions are as follows.
\begin{itemize}
    \item We identify a key limitation in existing ad-placement, diffusion-based relighting, and object-insertion methods: their inability to maintain lighting consistency for inserted ad banners.
    \item We introduce AD-Relight, a novel three-stage, training-free framework for relighting Photoshop-generated ad banners. Our method first aligns the banner’s shading with the target region, then employs a \textit{differential probing strategy} to extract region-specific illumination priors from a pre-trained lighting-aware diffusion model, and finally uses this prior to relight the banner. We provide a principled formulation and analysis of this differential illumination feature.
    \item We perform an extensive evaluation across diverse indoor lighting conditions and banners to ensure robustness. We conduct a user study to capture human perception of AD-Relight’s outputs.
\end{itemize}
\section{Related Work}

\subsection{End-to-End Ad Placement Methods}
\cite{akgul2020cloud} proposes a server-side dynamic ad insertion platform tailored for live streaming. \cite{efimova2022advertisement} presents a technique for replacing advertisements in videos using scene recognition and pattern matching. \cite{wang2008automatic} introduces an automatic editing framework for sports videos. The framework leverages computer vision techniques to detect key game events and camera transitions. \cite{rawat2019mid} proposes an emotion-aware ad placement method using multi-modal analysis.

End-to-end methods effectively predict the optimal regions and timestamps for ad placement to maximize user attention. However, they primarily focus on vertically aligned, portrait-style ads and often overlook the lighting information in the original scene. The original region exhibits shading and light gradients influenced by multiple light sources in the scene, whereas a newly inserted ad banner lacks these details. Ensuring seamless integration therefore requires a framework that relights the new banner to match the original lighting.
\subsection{Diffusion-based Object Insertion in Natural Scenes}
Several works extend diffusion models for photorealistic object insertion into user-specified regions. ObjectStitch \cite{song2023objectstitch} trains an adapter to generate CLIP-aligned embeddings for insertion, while TF-ICON \cite{lu2023tf} and \cite{tewel2024add} introduce training-free approaches that modify the reverse diffusion process to ensure spatial alignment and appearance consistency. AnyDoor \cite{chen2024anydoor} provides zero-shot object customization using a CLIP-based latent encoder to preserve object identity. Although diffusion-based insertion methods produce realistic, well-blended results, they often struggle to strictly preserve the identity and structure of inserted objects, an essential requirement for ad-banner placement.

\subsection{Object Relighting}
Several existing works use diffusion models or deep learning architectures, often combined with environment maps, to relight foreground objects based on background illumination. \cite{ren2024relightful} proposes a diffusion-based relighting model for indoor and outdoor scenes, integrating a physically based illumination model with a denoising network trained on a large set of relit examples. SwitchLight \cite{kim2024switchlight} co-designs a physics-driven architecture with a dedicated pretraining framework for human relighting. \cite{jin2024neural} explores neural rendering with volumetric representations, and IC-Light \cite{zhang2025scaling} uses diffusion for large-scale illumination harmonization.

While these models achieve strong results for relighting single objects via text or background conditioning, they face several limitations in banner relighting :
(i) some are not publicly available \cite{kim2024switchlight, ren2024relightful};
(ii) most are trained specifically for portrait-style, single-object relighting \cite{jin2024neural, fortier2024spotlight, zhang2025scaling}, making floor-level ad banners far outside their training distribution;
(iii) as shown in Fig. \ref{fig:intro}, when applied to such banners, they fail to maintain lighting consistency with the original scene.

\section{Proposed Method}
In this section, we introduce a framework for relighting a custom Photoshop-generated ad banner to ensure it reflects lighting behavior consistent with the original scene. Diffusion-based object relighting models effectively relight portrait-style objects conditioned on background lighting. However, they fail when applied to banners. To address this, we propose a framework that leverages the lighting understanding of a re-lighting model without requiring additional training. AD-Relight uses a publicly available model, IC-Light \cite{zhang2025scaling}, as it's re-lighting backbone. It is trained on over 10 million images with diverse lighting conditions, making it a suitable choice. Our framework consists of three main stages, which we describe in detail in the subsequent subsections.
\subsection{Stage 1: Shade Alignment}
\begin{figure*}[t]
    \centering
    \includegraphics[width=\textwidth, height = 0.30\textheight]{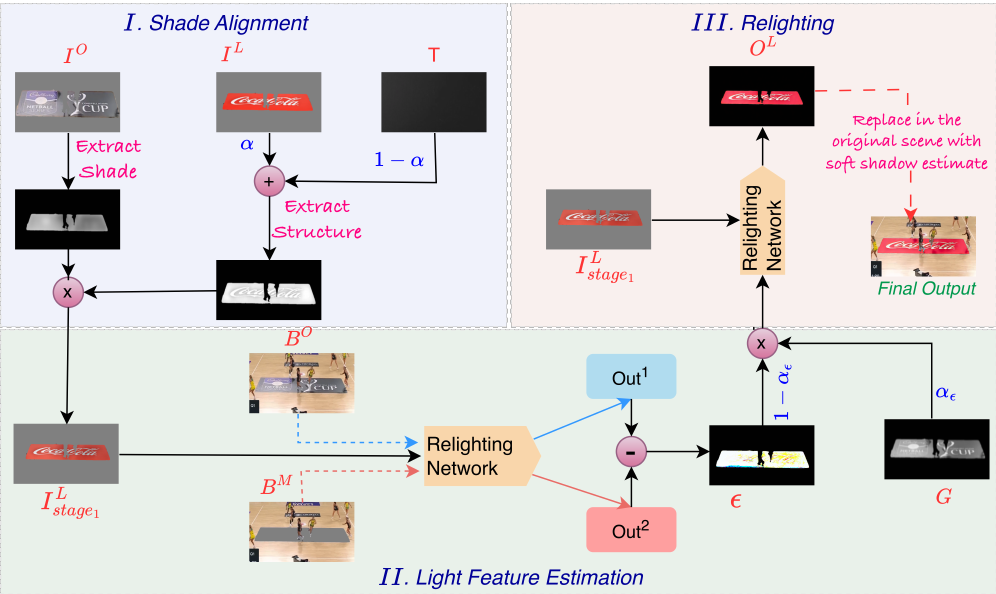}
    \caption{Overview of AD-Relight. In Stage 1, we align the shading characteristics of the input region with the custom banner. In Stage 2, we extract a region-specific lighting prior using a differential probing strategy on a pretrained relighting model. In Stage 3, this prior is used to relight the inserted banner, ensuring consistency with the scene illumination.}
    \label{fig:fig_3}
    \vspace{-0.5cm}
\end{figure*}
We provide an overview of Stage 1 in Fig. \ref{fig:fig_3}. Our method begins by extracting the input region using the Segment Anything Model (SAM) \cite{kirillov2023segment}, denoted as $I^{O}$. We pre-process the Photoshop-generated custom logo $I^{L}$ to match the lighting shade of the input region $I^{O}$. Additionally, we observe that ad banners in real-world scenes are typically printed on textured materials. To simulate this, we utilize a pre-trained diffusion model \cite{labs2025flux1kontextflowmatching} to generate a texture map $T$, which is alpha-blended with $I^{L}$ as follows: $I^{L} = (1-\alpha) I^{L} + \alpha T$, where $0 \leq \alpha \leq 1$. In our experiments, we set $\alpha = 0.3$.

Next, to estimate the shading map $S^O$ of the original region, we draw inspiration from Retinex theory \cite{land1971lightness, rother2011recovering}, which decomposes an image into illuminance and reflectance components. We approximate this decomposition by assuming that shading varies smoothly and can be captured by low-frequency components. Specifically, we factorize the illuminance channel of $I^O$ ($I^O_{\text{ill}}$) into $S^O_t$ and $S^O$. To obtain the structural information of the custom ad banner, we apply the same factorization to the illuminance channel of the ad banner ($I^L_{\text{ill}}$) as follows:
\begin{equation}
    S^O=GaussianFilter^K(I^O_{ill}), \quad S^O_t = \frac{I^O_{ill}}{S^O}
\end{equation}
\begin{equation}
    S^L=GaussianFilter^K(I^L_{ill}), \quad S^L_t = \frac{I^L_{ill}}{S^L}
\end{equation}
$$\implies I^O_{ill}=S^OS^O_t, \quad I^L_{ill}=S^LS^L_t$$

Here, $GaussianFilter$ is a smoothing filter of size $K$ that extracts the low-frequency shading component from the illuminance channel. This approximation is well-suited for planar surfaces where illumination changes are spatially smooth. $S^O_t$ and $S^L_t$ capture the structural information of $I^O$ and $I^L$, respectively, while $S^O$ and $S^L$ represent their corresponding shading components. We then transfer the shading of the original region to the banner by applying $S^O$ to the structure of $I^L$ as follows: $I^L_{ill} = S^O S^L_t$, ensuring that the custom logo inherits the shading characteristics of the original region.
\subsection{Stage 2: Light Feature Estimation}

Diffusion-based relighting models often fail on inserted ad banners, particularly when placed on horizontal surfaces, where they tend to reproduce floor color instead of scene-consistent illumination. Despite this limitation, we observe that such models encode useful illumination priors due to their training on large-scale relighting datasets. We therefore use IC-Light~\cite{zhang2025scaling} as a lighting backbone and probe it to extract region-specific illumination cues.

IC-Light enforces consistent light transport during training, encouraging approximately linear responses to illumination variations. While the model is not strictly linear, prior work \cite{zhang2025scaling} suggests that its outputs exhibit locally linear behavior with respect to illumination changes. We leverage this property in a differential manner.

To isolate the illumination contribution of the target region $I^O$, we construct two inputs. In the first, the full frame $B^O$ is used as the background. In the second, we remove $I^O$ to obtain a masked background $B^M$. Passing both inputs through the relighting backbone yields outputs $Out^1$ and $Out^2$, respectively.

Let $\phi$ denote the parameters of the backbone. The model implicitly encodes a light transport operator $T_\phi$ and an illumination representation $L_\phi$, such that:
$$Out^1 \approx T_\phi L_\phi, \quad Out^2 \approx T_\phi L^{M}_\phi$$
Assuming approximate linearity in illumination space, we write:
\begin{equation}
    L^{M}_\phi \approx L_\phi - L^{O}_\phi
    \label{eqn:lm}
\end{equation}
where $L^{O}_\phi$ denotes the illumination contribution of region $I^O$.

We define a differential illumination feature:
\begin{equation}
    \epsilon = Out^1 - Out^2
    \label{eqn:epsilon}
\end{equation}
Substituting Eq.~\ref{eqn:lm}, we obtain:
$$\epsilon \approx T_\phi L_\phi - T_\phi (L_\phi - L^{O}_\phi) \approx T_\phi L^{O}_\phi$$

Thus, $\epsilon \in \mathbb{R}^{H \times W \times 3}$ can be interpreted as a spatially aligned illumination residual in RGB space that captures the relative lighting contribution of $I^O$ under the model prior. While $\epsilon$ may contain minor appearance leakage due to imperfect disentanglement, it serves as an effective proxy for region-specific illumination. This differential probing allows us to extract reusable lighting priors from a pretrained relighting model without additional training.
\begin{figure*}[t] \centering \includegraphics[width=\textwidth]{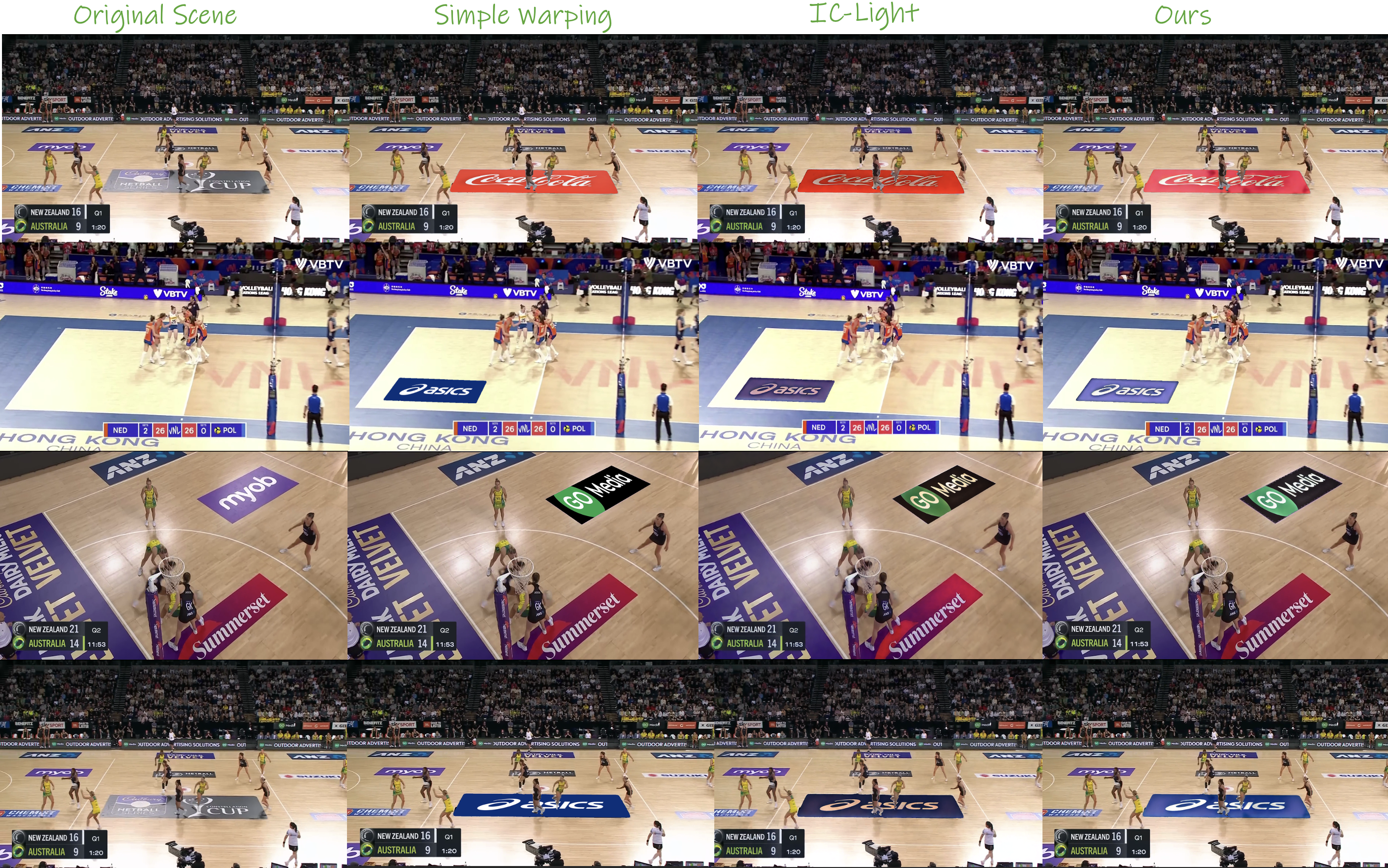} \caption{Qualitative comparison of AD-Relight with existing methods. Our approach produces lighting that is more consistent with the scene, particularly in challenging cases with strong illumination gradients and floor-pasted banners, resulting in improved visual integration.} \label{fig:fig_5}
\vspace{-15pt}
\end{figure*}
\subsection{Stage 3: Relighting the Custom Logo}
From Stage 2, we obtain the lighting feature $\epsilon$. Our objective is to ensure that our relighting backbone considers only the lighting information corresponding to the original region $I^O$ while relighting the custom logo $I^L$. To achieve this, as shown in Fig. \ref{fig:fig_3}, we pass a weighted combination of $\epsilon$ and an initial light gradient $G$ as the background input, denoted as $B^{\epsilon}$.

The initial light gradient $G$ is obtained by applying a Gaussian filter to the illuminance channel of $I^O$ ($I^O_{\text{ill}}$), capturing low-frequency illumination structure. Combining $\epsilon$ with $G$ suppresses high-frequency artifacts in $\epsilon$ while preserving global lighting trends:
$$G = GaussianFilter^{K^{\prime}}(I^O_{\text{ill}}), \quad B^{\epsilon} = \alpha_\epsilon G + (1- \alpha_\epsilon) \epsilon$$
\begin{equation}
    O^L = ICLight(B^\epsilon, I^L)
\end{equation}
Here, $K^{\prime} < K$, and $\alpha_\epsilon$ balances global illumination ($G$) and differential lighting ($\epsilon$).

To further enhance realism, we incorporate soft shadow estimation. We compute an adaptive threshold $Thr$ using Otsu's method on $I^O_{\text{ill}}$, enabling robustness to varying illumination conditions. We obtain a shadow map by identifying low-illuminance regions and define a continuous shadow attenuation factor:
\begin{equation}
    df(x, y) = \min\left(1, \frac{I^{O}_{\text{ill}}(x, y)}{Thr}\right)
\end{equation}
This formulation produces a smooth attenuation that preserves relative intensity differences while avoiding hard transitions. To reduce sensitivity to texture noise, the illuminance map is pre-smoothed via Gaussian filtering.

We apply this attenuation to the relit logo and blend it with the original output using $\alpha_s = 0.2$:
$$I^{O^{L}}_{\text{ill}} = \alpha_s I^{O^{L}}_{\text{ill}} + (1 - \alpha_s) I^{O^{L}}_{\text{ill}} \cdot df$$
\section{Results}
\subsection{Qualitative Analysis}
In this work, we investigate single-frame relighting. Fig. \ref{fig:fig_5} presents a qualitative comparison of our method against two baselines: simple warping, commonly used in ad-placement methods, and IC-Light \cite{zhang2025scaling}. As shown, simple warping fails to incorporate lighting gradients and uniformly maps the logo to the region, disregarding scene illumination. IC-Light, a state-of-the-art diffusion-based relighting method, performs well in standard settings but struggles in challenging cases where banners are placed on horizontal surfaces with complex lighting. In such scenarios, it tends to bias the logo’s appearance toward surrounding textures rather than accurately reproducing scene-consistent illumination.

Our benchmark includes a diverse set of scenes with varying lighting conditions, viewpoints, and materials, including both indoor and outdoor examples. While existing methods perform reasonably well on simpler configurations, we focus on more challenging cases involving floor-pasted banners with strong light gradients, soft shadows, and non-uniform illumination. We select representative examples to highlight these cases. Row 1 presents a scene with a strong light gradient and soft shadows. Row 2 shows a glossy wooden floor under intense lighting. Row 3 illustrates a top-view camera angle. Row 4 uses the same scene as Row 1 with a different colored banner. Additional qualitative results are provided in the supplementary material.

\subsection{Comparative Analysis}
\noindent \textbf{Quantitative comparison.} We evaluate our approach against four baselines: \textbf{TF-ICON} \cite{lu2023tf}, a training-free DDIM inversion-based object insertion method, \textbf{IC-Light} \cite{zhang2025scaling}, and \textbf{SpotLight} \cite{fortier2024spotlight}, both designed for background-conditioned relighting, and \textbf{simple warping}, a widely adopted technique in ad-placement pipelines. 

Our evaluation is conducted on a benchmark constructed following prior protocols \cite{frenkel2024implicit,liu2024unziplora}, consisting of 560 cases generated from 7 logos and 80 frames. The benchmark is designed to include diverse lighting conditions and challenging placements, particularly cases with strong illumination variation, complex surface textures, and horizontal banner placement.

\begin{table}[h]
\vspace{-0.3cm}
\caption{Quantitative comparison. Best results are shown in green (and bold), second-best in red.}
\centering
\begin{tabular}{lccc}
\hline
\textbf{Method} & \textbf{SSIM $\uparrow$} & \textbf{LPIPS $\downarrow$} & \textbf{ILL-SIM $\uparrow$} \\
\hline
Simple Warping & 0.89 & 0.12 & {\cellcolor{red!15}0.82} \\
TF-ICON        & 0.45 & 0.57 & 0.62\\
IC-Light       & 0.91 & {\cellcolor{red!15}0.07} & 0.77\\
SpotLight      & {\cellcolor{red!15}0.92} & 0.09 & 0.79\\
\textbf{Ours}  & {\cellcolor{green!15}\textbf{0.95}} & {\cellcolor{green!15}\textbf{0.03}} & {\cellcolor{green!15}\textbf{0.92}}\\
\hline
\end{tabular}
\label{tab:table1}
\end{table}
\vspace{-0.1cm}

Following \cite{zhang2025scaling,jin2024neural,fortier2024spotlight}, we evaluate our approach using LPIPS and SSIM scores for the relighted region. Table \ref{tab:table1} presents a comparison with existing baselines. We also report the cosine similarity (ILL-SIM) between the illuminance of the input region and the relighted banner, which reflects the consistency of illumination transfer.

\noindent \textbf{LLM evaluation.} Following recent generative evaluation protocols \cite{ouyang2025k}, 
\begin{table}[h]

\caption{LLM evaluation. Values indicate the percentage of cases in which AD-Relight is preferred over competing methods across different lighting-related criteria.}
\centering
\begin{tabular}{l|ccc}
\hline
\multicolumn{4}{c}{\textbf{GPT-4o Evaluation: \% preference of Ad-Relight over:}} \\
\hline
\textbf{Criteria} & Simple Warping & IC-Light & SpotLight \\
\hline
Light Gradient   & 100\%  & 97.6\% & 95.2\% \\
Light Consistency & 97.6\% & 95.2\% & 92.8\% \\
Scene Realism     & 90.4\% & 92.8\% & 85.7\% \\
\hline
\end{tabular}
\label{tab:tab2}
\end{table}
we conduct an LLM-based assessment where GPT-4o ranks outputs from IC-Light, simple warping, SpotLight, and our method based on light gradient, light consistency, and realism. Table \ref{tab:tab2} shows that GPT consistently prefers our method across all criteria, with particularly strong preference in cases involving complex illumination. Details of prompts and ILL-SIM computation are provided in the supplementary material.

\noindent \textbf{User study.} To further evaluate our method, we conducted a user study to assess human perception of the generated results. Four questionnaire variants were prepared, each containing nine scenarios with two unlabeled relighted options. 
\begin{figure}[h] 
    \centering \includegraphics[width=\columnwidth, keepaspectratio]{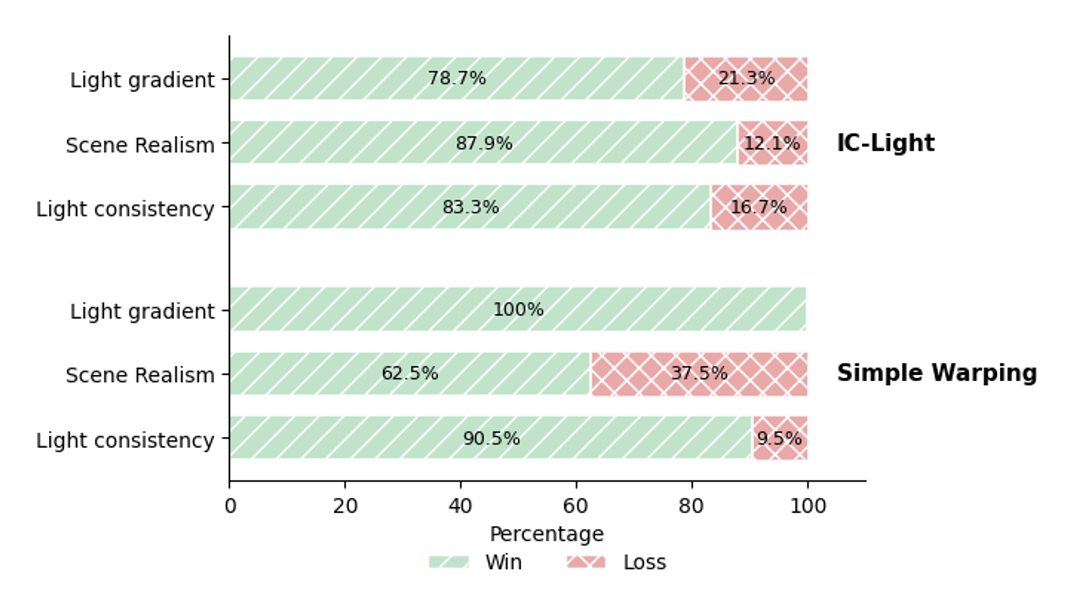} \caption{\textbf{Human User Study.} The green portion of each bar indicates the percentage of responses in which participants preferred our results over the corresponding baseline.}  \label{fig:fig_graph} 
    \vspace{-10pt}
\end{figure}
A total of 25 participants provided 324 responses. The results, shown in Fig. \ref{fig:fig_graph}, indicate that users consistently preferred our outputs, particularly in challenging lighting scenarios.
\
\subsection{Additional analysis}
The key hyperparameters for AD-Relight are $K=99$ and $\alpha=0.3$ in Stage 1, and $\alpha_\epsilon=0.4$, $K^{\prime}=21$, and $\alpha_s=0.2$ in Stage 3. In this section, we present ablation studies to justify these choices.

We compare results using two variants of hyperparameter settings. In \textbf{M1}, we use $\alpha = 0.2$, $\alpha_s = 0.1$, $\alpha_\epsilon = 0.2$, and $K^{\prime} = 71$. In \textbf{M2}, we use $\alpha = 0.4$, $\alpha_s = 0.3$, $\alpha_\epsilon = 0.5$, and $K^{\prime} = 15$. In both cases, we consider values above and below the final configuration. To further evaluate the contribution of each component, we compare against \textbf{M3} (without $G$), \textbf{M4} (without shade alignment), and \textbf{M5} (without $\epsilon$). Results are reported in Table \ref{tab:table3}.
\begin{table}[t] 
    \caption{Ablation results.} 
    \centering 
    \begin{tabular}{lccc} \hline \textbf{Method} & \textbf{SSIM $\uparrow$} & \textbf{LPIPS $\downarrow$} & \textbf{ILL-SIM $\uparrow$} \\ \hline M1 & 0.92 & 0.03 & 0.85 \\ M2 & 0.91 & 0.05 & 0.87 \\ M3 & 0.89 & 0.07 & 0.83 \\ M4 & 0.86 & 0.05 & 0.80 \\ M5 & 0.80 & 0.10 & 0.85 \\ \rowcolor{gray!30} \textbf{Ours} & \textbf{0.95} & \textbf{0.03} & \textbf{0.92} \\ \hline 
    \end{tabular} 
    \label{tab:table3} 
    \vspace{-0.3cm} 
    \end{table}
    \begin{table}[h] 
    \caption{The user study shows that participants consistently prefer our full framework over its variants.} 
    \centering 
    \begin{tabular}{l|ccc} 
    \hline \multicolumn{4}{c}{\textbf{Preference (\%) for Ad-Relight over:}} \\ \hline \textbf{Criteria} & M3 & M4 & M5 \\ \hline Light Gradient & 55.0\% & 80.5\% & 63.8\% \\ Light Consistency & 55.0\% & 69.4\% & 87.5\% \\ Scene Realism & 61.1\% & 52.0\% & 72.2\% \\ \hline 
    \end{tabular} 
    \label{tab:tab4} 
    \end{table}
    
\begin{figure}[h] 
    \centering 
    \includegraphics[width=\columnwidth]{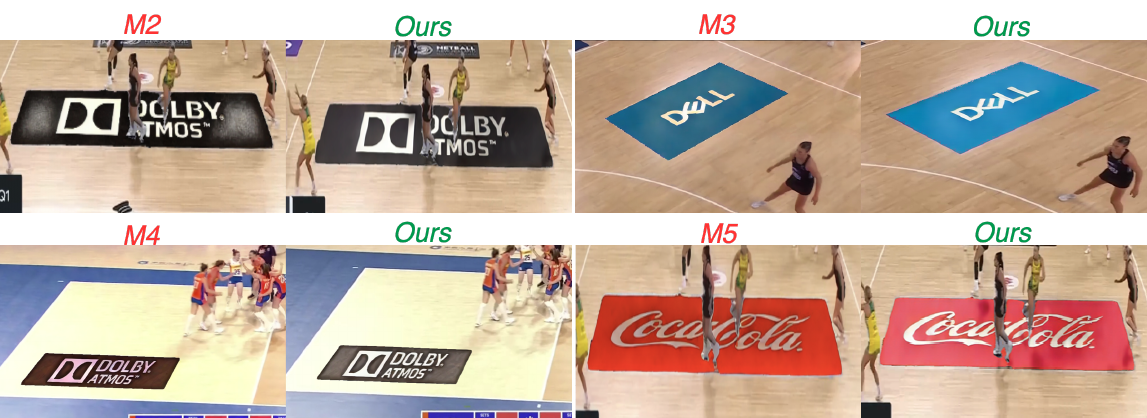} 
    \caption{Qualitative ablations} 
    \label{fig:ablation} 
    \vspace{-0.2cm} 
\end{figure}
\noindent \textbf{Limitations.} In this work, we focus on single-frame ad banner relighting. Our approach relies on the backbone’s implicit disentanglement of illumination, and performance may degrade when this assumption does not hold. In the supplementary, we discuss the extension of our work to temporally coherent video banner relighting, and its limitations
along with additional qualitative analysis, and details of user study setup.
\section{Conclusion}
In this work, we present AD-Relight, a novel training-free framework for relighting custom ad banners to ensure consistency with the lighting conditions of the original scene. Our findings reveal that while ad placement methods effectively identify suitable regions for placing ads, they fail to adjust lighting for realistic integration. Object relighting models, though effective for portrait-type objects, produce inconsistent lighting when applied to ad banners. To overcome these limitations, we propose a multi-stage framework that adapts a state-of-the-art publicly available diffusion-based relighting model for banner relighting without additional training. We validate AD-Relight across multiple scenes with diverse lighting conditions and test it on various ad banners with different structures and colors to support it's effectiveness. We also identify clear limitations that motivate future work, particularly in temporal coherence.
\bibliographystyle{IEEEbib}
\bibliography{ref}
\end{document}